\documentclass[letterpaper, 10 pt, conference]{ieeeconf} 
\IEEEoverridecommandlockouts                              

\usepackage{amsmath}
\usepackage{amssymb}
\usepackage{xcolor}
\usepackage{hyperref}
\usepackage{graphicx}
\usepackage{subcaption}

\definecolor{custompink}{RGB}{239,118,186} 
\hypersetup{
    colorlinks=true,
    linkcolor=blue,
    citecolor=green,
    filecolor=magenta,
    urlcolor=custompink
}

\title{\LARGE \bf Optimal Actuator Attacks on Autonomous Vehicles Using Reinforcement Learning}

\author{Pengyu Wang$^{\dag, 1,2}$, Jialu Li$^{\dag, 1}$ and Ling Shi$^{1}$, \textit{Fellow, IEEE}
\thanks{$^{\dag}$ Equal contribution.}
\thanks{$^{1}$Pengyu Wang, Jialu Li and Ling Shi are with the Department of Electronic and Computer Engineering, Hong Kong University of Science and Technology, Hong Kong SAR. {\tt\small \{pwangat, jlikr\}@connect.ust.hk, eesling@ust.hk}}
\thanks{$^{2}$Pengyu Wang is also with the Shenzhen Key Laboratory of Robotics Perception and Intelligence and the Department of Electronic and Electrical Engineering, Southern University of Science and Technology, Shenzhen, China.}
}

\begin{document}

\maketitle
\thispagestyle{empty}
\pagestyle{empty}

\begin{abstract}
With the increasing prevalence of autonomous vehicles (AVs), their vulnerability to various types of attacks has grown, presenting significant security challenges. In this paper, we propose a reinforcement learning (RL)-based approach for designing optimal stealthy integrity attacks on AV actuators. We also analyze the limitations of state-of-the-art RL-based secure controllers developed to counter such attacks. Through extensive simulation experiments, we demonstrate the effectiveness and efficiency of our proposed method.
\end{abstract}

\section{INTRODUCTION}
\begin{figure}[t!]
    \centering
    \hfill 
    \begin{subfigure}[b]{0.49\columnwidth}
        \includegraphics[width=\linewidth]{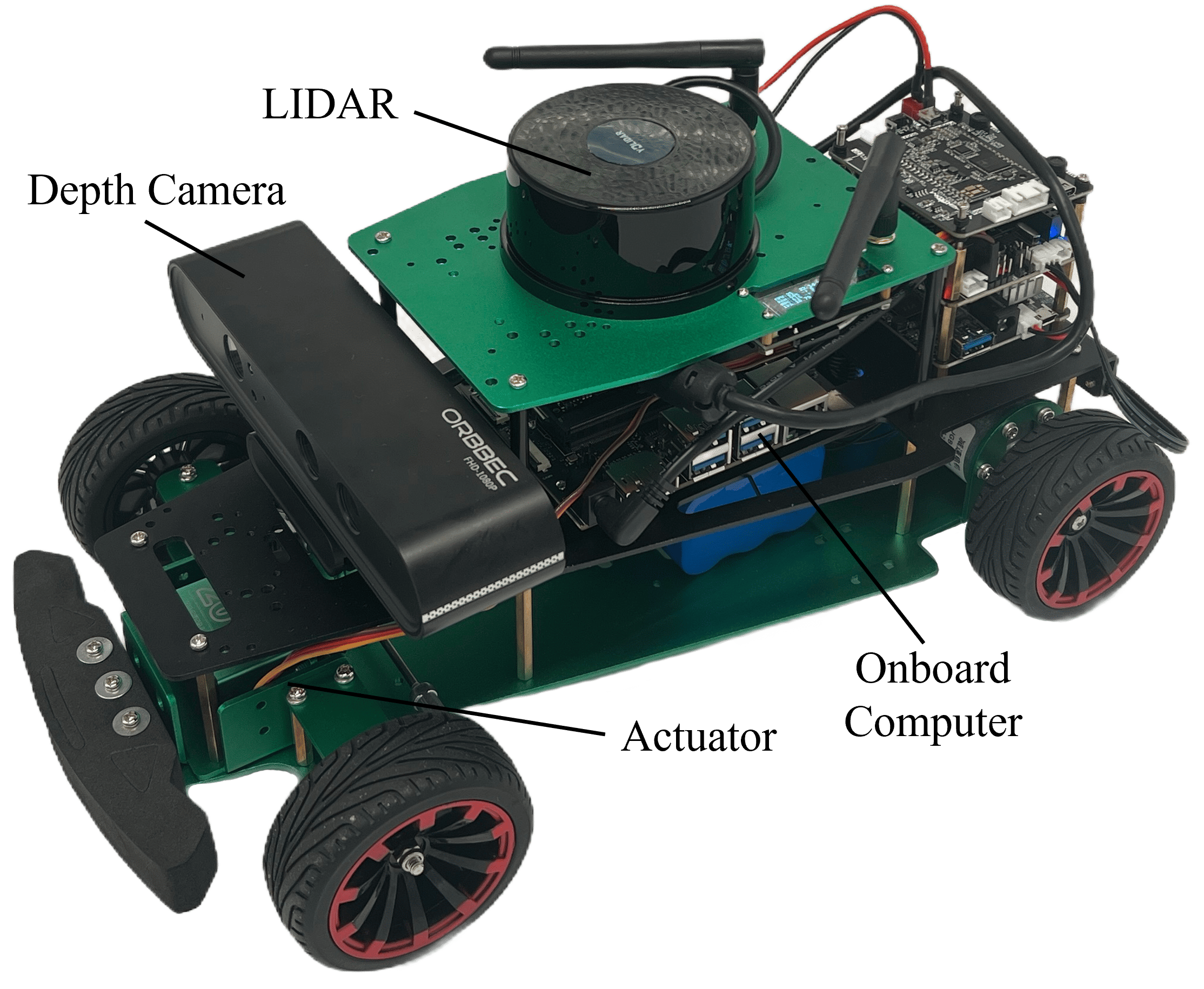}
        \caption{Autonomous vehicle.}
        \label{fig:av}
    \end{subfigure}
    \hfill
    \begin{subfigure}[b]{0.49\columnwidth}
        \includegraphics[width=\linewidth]{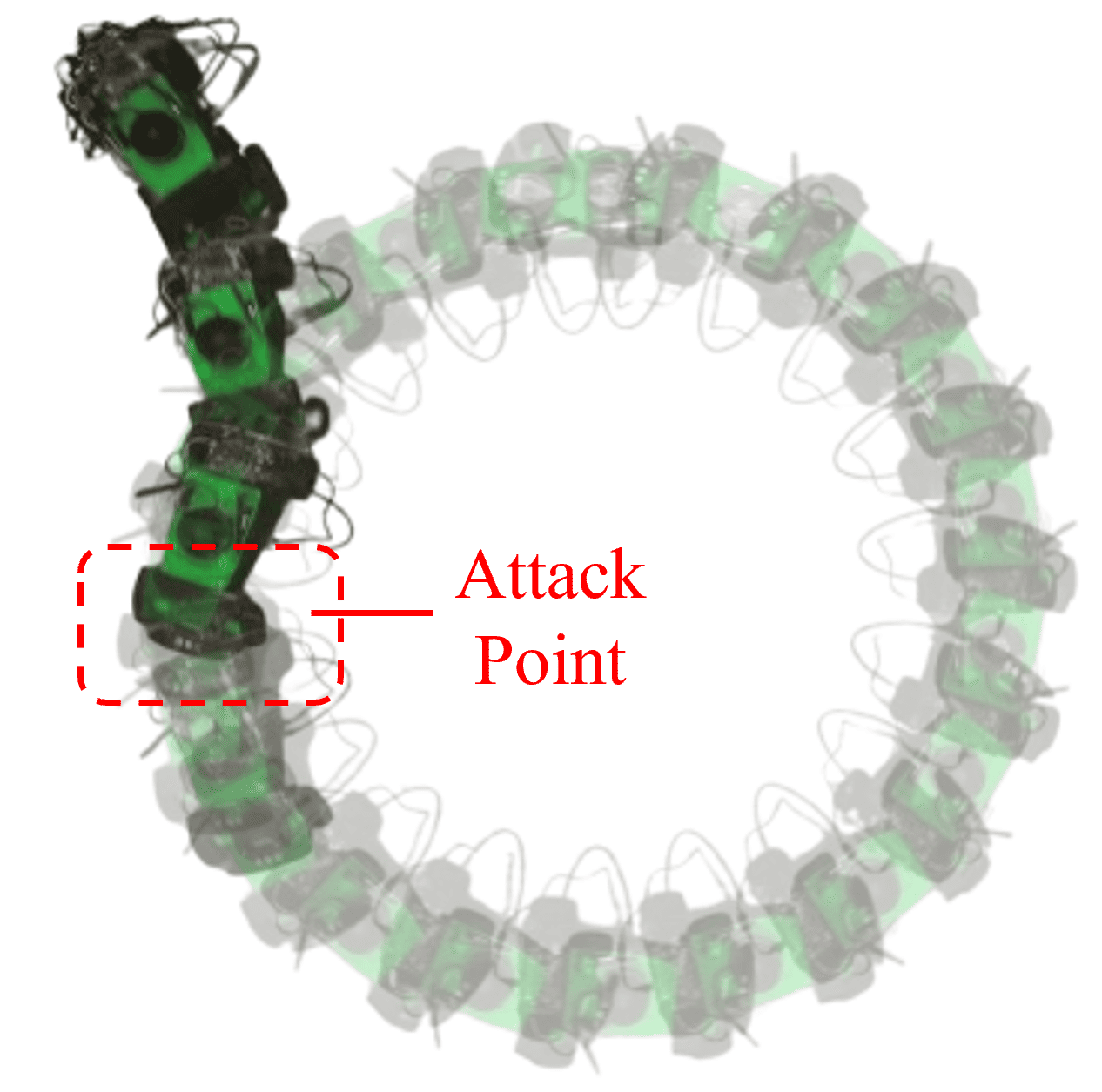}
        \caption{AV under attack.}
        \label{fig:attack}
    \end{subfigure}
    \caption{AV attacked while executing a circular trajectory.}
    \label{fig:av_attack}
\end{figure}

With the emergence of autonomous vehicles featuring advanced sensors and accurate actuators, their application has expanded across a wide range of scenarios~\cite{wang2024apf,wang2025miner}. Recently, there has been a growing focus on the security and safety issues associated with these vehicles. Due to the high reliance of autonomous vehicles on software and communication systems, they are vulnerable to different types of attacks, as shown in Fig.~\ref{fig:av_attack}, which may lead to severe accidents~\cite{ahmad2024comprehensive}. Attacks on autonomous vehicles are typically categorized into those targeting actuators and those targeting sensors. Compared to sensor attacks, actuator attacks are more challenging to detect since it does not directly affect the observations~\cite{yang2023lasso}. Previous research has typically focused on several types of actuator attacks, including denial-of-service (DoS)~\cite{huang2022learning}, false data injection attacks (FDI)~\cite{wu2023secure}, and replay attacks. FDI attacks are more common and easier to be implemented than replay attacks, while being more covert and deceptive than DoS attacks, and are therefore the focus of this paper.

In~\cite{wu2018optimal}, the authors analyzed the dynamic response of a system under optimal switching data injection attacks. Gao \textit{et al.}~\cite{gao2021class} designed an optimal mixed data injection attack strategy, combining false data and its derivative, to minimize a quadratic cost and degrade system performance. However, both works are limited to linear systems. In the most recent work, Wu \textit{et al.}~\cite{wu2023secure}, for the first time, focused on the optimal actuator FDI attack design and the corresponding optimal secure controller for nonlinear autonomous vehicles using reinforcement learning. While the paper advances our understanding, yet there are two limitations. First, it lacks a focus on the stealthiness of the attacks, which is crucial given that modern autonomous vehicle systems are equipped with advanced attack detectors. Second, the design of their secure controller is based on specific FDI attack and training data obtained through RL, limiting its generalization to different types of attacks. These limitations motivate our research.

\section{METHODOLOGY}
\subsection{System Model}
Consider the following model of an autonomous vehicle's state and measurement:
\begin{equation}\label{eq:system_model}
\begin{aligned}
\mathbf{x}_{k+1} &= f(\mathbf{x}_{k}, u_{k}, w_{k}) = \begin{bmatrix}
v_{k} \cos \theta_{k} \\
v_{k} \sin \theta_{k} \\
\frac{v}{L} \tan \phi_{k}
\end{bmatrix} + w_{k}, \\
\mathbf{y}_{k+1} &= g(\mathbf{x}_{k}, q_{k}) = \left[\begin{array}{c}
\sqrt{\left(p_{x}-x_{k}\right)^{2}+\left(p_{y}-y_{k}\right)^{2}} \\
\arctan \left(\frac{p_{y}-y_{k}}{p_{x}-x_{k}}\right)-\theta_{k}
\end{array}\right] + q_{k},
\end{aligned}
\end{equation}
where $\mathbf{x}_{k} = [x_{k},y_{k},\theta_{k}] \in \mathbb{R}^3$ is the state, $u_{k} = [v_{k}, \phi_{k}] \in \mathbb{R}^2$ is the control input, $\mathbf{y}_{k} \in \mathbb{R}^2$ is the range and bearing measurement, $p = [p_x, p_y]$ is the position of landmark detected by the range sensor, $L$ is the wheelbase, $w_{k} \sim N(0, Q_{k}) \in \mathbb{R}^2$ and $q_{k} \sim N(0, R_{k}) \in \mathbb{R}^2$ are zero-mean Gaussian noises.

We consider there are attackers on the vehicles' actuators. The attack model is:
\begin{align}\label{eq:fdi}
    \mathbf{x}_{k+1} &= f(\mathbf{x}_{k}, u_{k} + d_{k}, w_{k}),
\end{align}
where $d_{k} = [v_d, \phi_d] \in \mathbb{R}^2$ is the injected false data.

\subsection{Problem Formulation}
For autonomous vehicles, the objective is typically to follow a reference trajectory $\mathbf{x}_{r,k+1} \in \mathbb{R}^3$, so the attacker's primary goal is to cause the vehicle to deviate from this trajectory, and we consider the cost function as $J_{t} = [\mathbf{x}_{k} - \mathbf{x}_{r,k}]^{T}Q [\mathbf{x}_{k} - \mathbf{x}_{r,k}]$. Additionally, the attacker aims to attack with minimal energy expenditure, and the cost function is $J_{e} = u_{k}^{T}R u_{k}$. Finally, the attacker seeks to maintain stealthiness against common residue-based attack detectors~\cite{yang2023lasso, liu2019secure, wang2024quadformer}, and thus the corresponding cost function is $J_{s} = \alpha (1-D_{k})$, where $\alpha$ is a scaling parameter, and $D_{k} = \left \{ 0, 1\right \} $ is the binary output of the detector. The final objective function is written as: 
\begin{equation}\label{eq:obj}
J = \lim_{N \to \infty} \frac{1}{N} \sum_{k=1}^{N} (J_{t} - J_{e} + J_{s}),
\end{equation}
where $N$ is the total attacked steps. 

Consider system (\ref{eq:system_model}) with possible attacks (\ref{eq:fdi}) acting on the actuator, design an optimal attack strategy such that (\ref{eq:obj}) is maximized. 

\subsection{Reinforcement Learning}
We model this process as a Markov decision process and define our reward function as follows:
\begin{equation}\label{eq:reward}
R_{k} = J_{t} - J_{e} + J_{s}.
\end{equation}
We leverage the off-policy Proximal Policy Optimization (PPO)~\cite{schulman2017proximal} and on-policy Soft Actor-Critic (SAC)~\cite{haarnoja2018soft} algorithm to solve the above non-linear and non-convex problem.

\section{EXPERIMENTS AND ANALYSIS}
\subsection{Experiment Setting}
The controller is adopted from~\cite{wang2015simultaneous} and an extended Kalman filter (EKF) is used for state estimation and residue generation. The vehicle trajectory tracking environment is built using OpenAI Gym. Following~\cite{yang2023lasso}, we design a dynamic $\chi ^{2}$ attack detector based on the prediction residue:
\begin{equation}
    r_{k} =\hat{\mathbf{x}}_{k \mid k}-\hat{\mathbf{x}}_{k \mid k-1},
\end{equation}
which is the difference between the state estimate $\hat{\mathbf{x}}_{k \mid k}$ and the controller prediction $\hat{\mathbf{x}}_{k \mid k-1}$. 

\subsection{Result Analysis}
\begin{figure}[h]
    \centering
    \hfill 
    \begin{subfigure}[b]{0.49\columnwidth}
        \includegraphics[width=\linewidth]{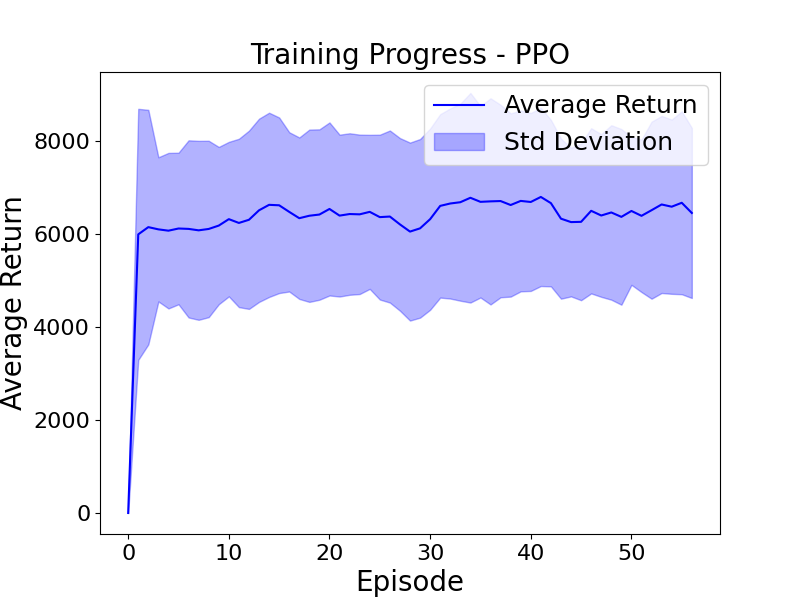}
        \caption{PPO result.}
        \label{fig:ppo}
    \end{subfigure}
    \hfill
    \begin{subfigure}[b]{0.49\columnwidth}
        \includegraphics[width=\linewidth]{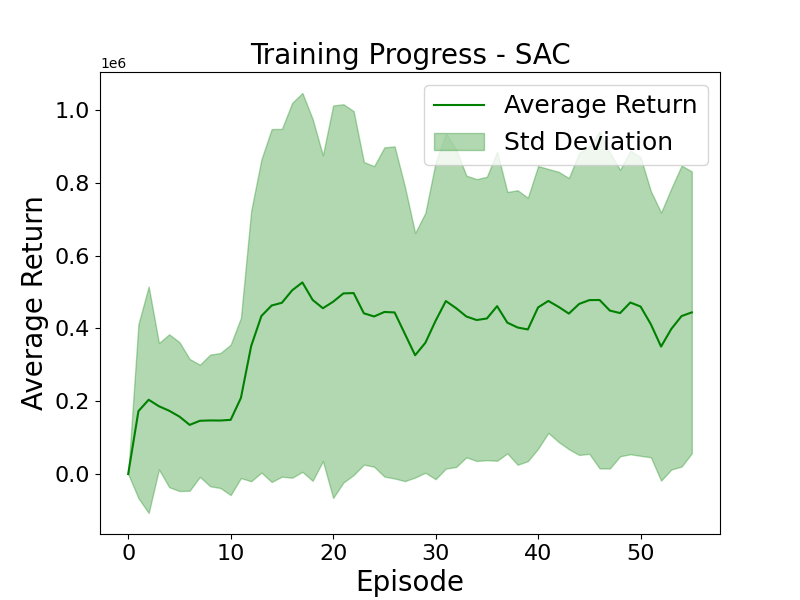}
        \caption{SAC result.}
        \label{fig:sac}
    \end{subfigure}
    \caption{Training results for two RL algorithms.}
    \label{fig:training}
\end{figure}

We trained PPO and SAC for 50 episodes until convergence, and the results are shown in Fig.~\ref{fig:training}. Compared to SAC, PPO is easier to train, more stable, and achieves better performance in our tasks.

We then deployed the model to carry out real attacks on the actuator and evaluated the performance in two different attack cycle scenarios, as shown in Tab.~\ref{tab:attack 1} and Tab.~\ref{tab:attack 2}. For evaluation indicators, the detector recall $\frac{\text{TP}}{\text{TP} + \text{FN}}$ reflects the stealthiness, where there is an attack but it is not detected. The energy consumption $\frac{1}{N} \sum_{k=1}^{N} J_{e}$ is the average energy consumed per attack step. The tracking error $\frac{1}{N} \sum_{k=1}^{N} J_{t}$ is the average deviation from the trajectory caused by the attacker to the AV in each attack step. The experimental results demonstrate that, compared to~\cite{wu2023secure}, our method not only optimizes tracking error and attack energy but also takes into account the stealthiness of the attack, making the attacker more harmful to AV.

\begin{table}[h]
    \caption{Results for Attack Scenario 1.}
    \begin{center}
    \renewcommand\arraystretch{1.2}  
    \resizebox{\columnwidth}{!}{     
        \begin{tabular}{c|ccc}
        \hline
            Long Attack & Detector Recall & Energy Consumption & Tracking Error \\ 
        \hline
            SAC & 0.0709 & 10007 & 3154  \\
            PPO & 0.0825 & 4079 & 1362 \\ 
        \hline
        \end{tabular}
    }
    \end{center}
    \label{tab:attack 1} 
\end{table}
\begin{table}[htbp]
    \caption{Results for Attack Scenario 2.}
    \begin{center}
    \renewcommand\arraystretch{1.2}  
    \resizebox{\columnwidth}{!}{     
        \begin{tabular}{c|ccc}
        \hline
            Short Attack & Detector Recall & Energy Consumption & Tracking Error \\ 
        \hline
            SAC & 0.1339 & 10005 & 3215 \\
            PPO & 0.1518 & 4450 & 1460 \\ 
        \hline
        \end{tabular}
    }
    \end{center}
    \label{tab:attack 2} 
\end{table}
\section{DISCUSSION}
In this paper, we propose a novel learning-based method to design optimal stealthy actuator attacks on autonomous vehicles. Extensive simulation experiments have demonstrated the performance of our algorithm. In addition, we identify two distinct issues with the current state-of-the-art learning-based secure countermeasure~\cite{wu2023secure}, which we propose as directions for future research. First, the secure controller in~\cite{wu2023secure} is designed specifically for its original RL-based attacker, meaning that the controller is limited by the attack strategies learned through RL. Its effectiveness against different attack strategies (e.g., variations in frequency, amplitude, or pattern) or even different types of attacks remains unknown, and the authors do not discuss these scenarios. Second, while~\cite{wu2023secure} provides a proof of stability for its secure controller, the expansion of the attack space, particularly in the case of sensor attacks, introduces challenges to the convergence of RL under different attack scenarios. In fact, for real-world defenders, the attacker's model is typically impossible to be fully known. Moreover, as discussed in this paper, attackers often exhibit highly stealthy characteristics, making the two issues raised above even more significant. In the future, we plan to deploy our algorithm in real-world autonomous vehicles and design a new secure controller to address the above two issues.

\section*{APPENDIX}

\subsection{Attack Detector Operation Process}

\begin{figure}[h]
    \centering
    \includegraphics[width=0.98\columnwidth]{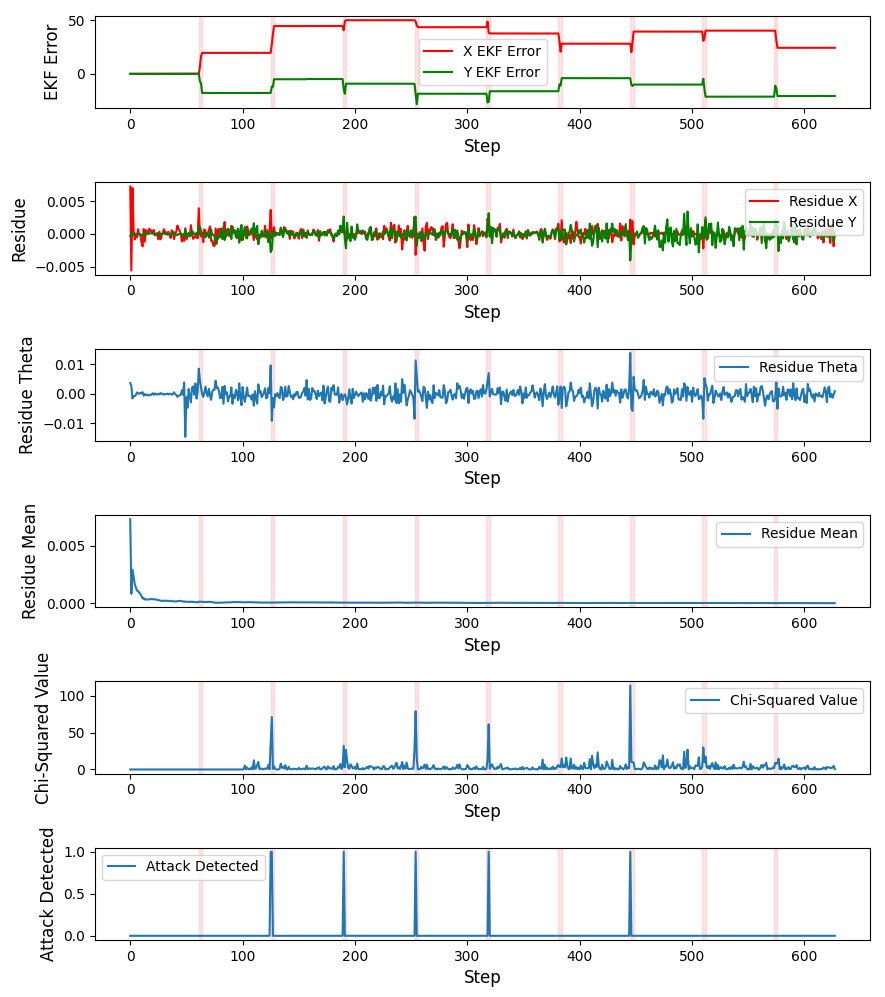} 
    \caption{Dynamic $\chi ^{2}$ attack detector operation process.}
    \label{fig:detector}
\end{figure}

We visualize the detection process using an advanced attack detector~\cite{yang2023lasso} when deploying our RL-based attacker. The light red areas indicate the attack periods, and the results show that the advanced residue-based attack detector misses many attack instances, which further validates the stealthiness of our method, as shown in Fig~\ref{fig:detector}.

\subsection{AV's Trajectory in Simulator}

We use a simulator to visualize the impact of our designed RL-based attacker on an autonomous vehicle. With good stealthiness, our method can cause the AV to deviate from its original trajectory through sparse attacks.

\subsubsection{Training Phase}
\begin{figure}[h]
    \centering
    \hfill 
    \begin{subfigure}[b]{0.49\columnwidth}
        \includegraphics[width=\linewidth]{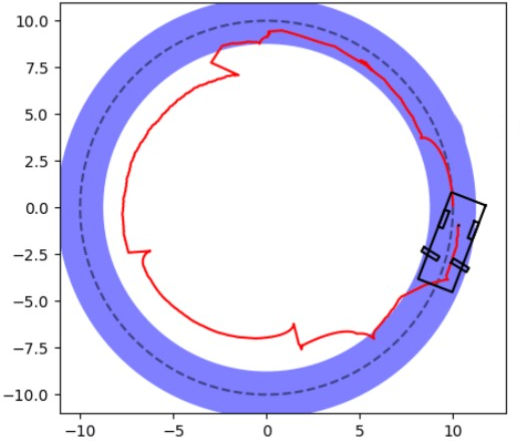}
        \caption{Episode 2.}
        \label{fig:training_1}
    \end{subfigure}
    \hfill
    \begin{subfigure}[b]{0.49\columnwidth}
        \includegraphics[width=\linewidth]{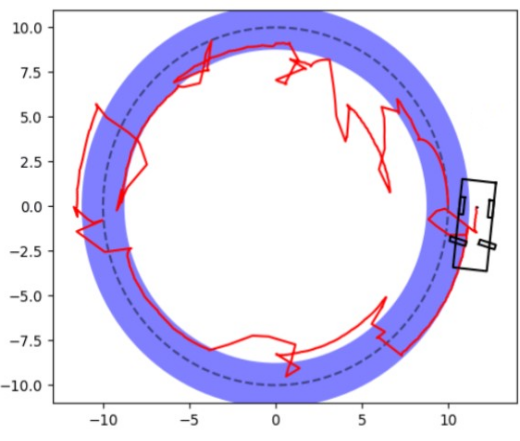}
        \caption{Episode 10.}
        \label{fig:training_2}
    \end{subfigure}
    \caption{Trajectory visualization in training phase.}
    \label{fig:training_traj}
\end{figure}

During the training phase, our designed attacker gradually learns better attack strategies as episodes progress, with the results visualized in the simulator as shown in Fig~\ref{fig:training_traj}. As training progresses, single-step attacks cause the autonomous vehicle to deviate much further from its original trajectory while maintaining stealthiness.

\subsubsection{Inference Phase}
\begin{figure}[h]
    \centering
    \hfill 
    \begin{subfigure}[b]{0.49\columnwidth}
        \includegraphics[width=\linewidth]{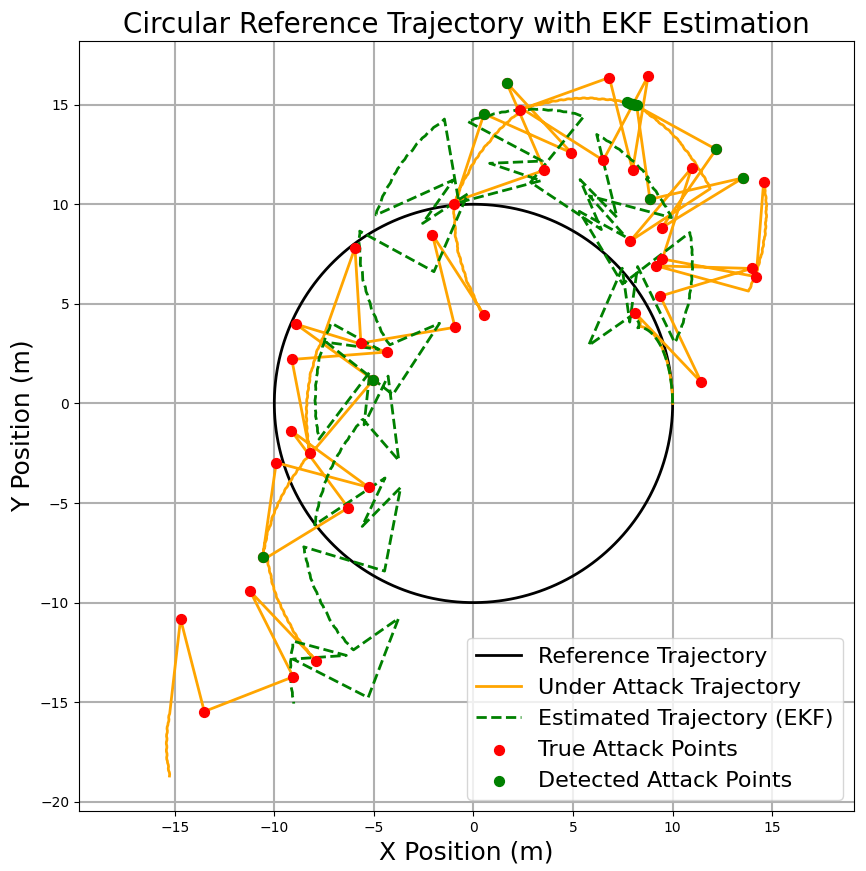}
        \caption{PPO result.}
        \label{fig:inference_1}
    \end{subfigure}
    \hfill
    \begin{subfigure}[b]{0.49\columnwidth}
        \includegraphics[width=\linewidth]{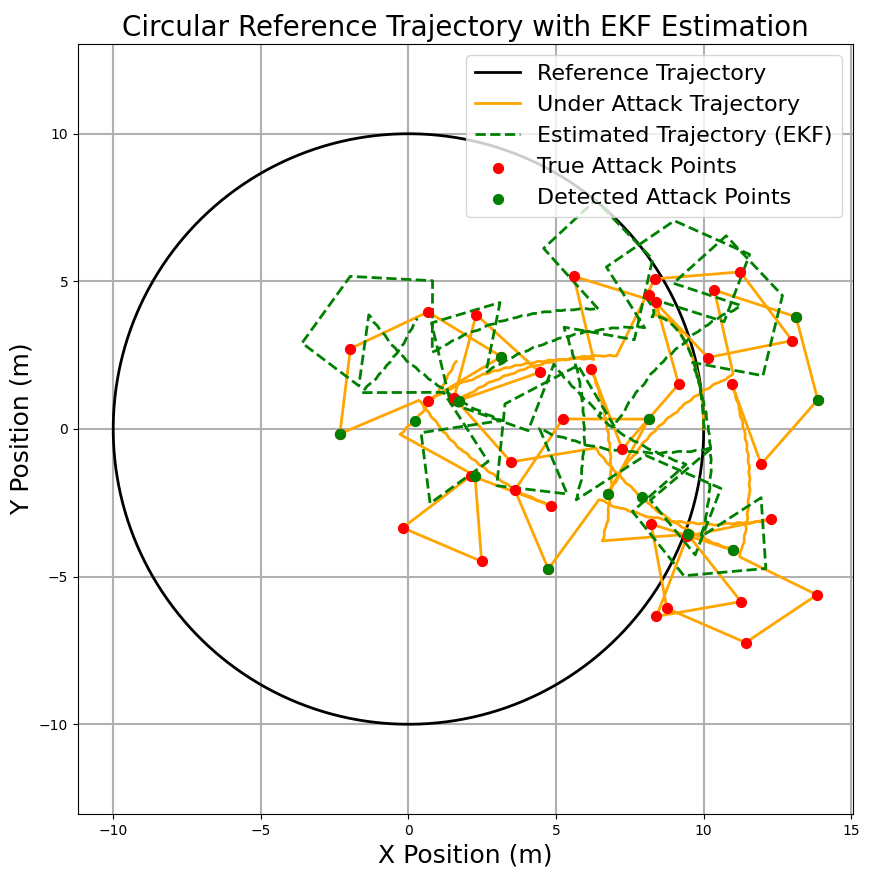}
        \caption{SAC result.}
        \label{fig:inference_2}
    \end{subfigure}
    \caption{Trajectory visualization in inference phase.}
    \label{fig:inference_traj}
\end{figure}

In the inference phase, our designed attacker causes the vehicle to deviate from its original trajectory, while the EKF effectively tracks the post-attack trajectory, indicating that the attack detector is functioning normally. The results shown in Fig.~\ref{fig:inference_traj} illustrate the actual effectiveness of our attacker. 

\bibliographystyle{ieeetr}
\bibliography{main}

\end{document}